# CPSOR-GCN: A Vehicle Trajectory Prediction Method Powered by Emotion and Cognitive Theory

*Lanyue Tang, Lishengsha Yue, (Member, IEEE), Jinghui Yuan, Jian Sun, Aohui Fu*

***Abstract*—Active safety systems on vehicles often face problems with false alarms. Most active safety systems predict the driver's trajectory with the assumption that the driver is always in a normal emotion, and then infer risks. However, the driver's trajectory uncertainty increases under abnormal emotions. This paper proposes a new trajectory prediction model: CPSOR-GCN, which predicts vehicle trajectories under abnormal emotions. At the physical level, the interaction features between vehicles are extracted by the physical GCN module. At the cognitive level, SOR cognitive theory is used as prior knowledge to build a Dynamic Bayesian Network (DBN) structure. The conditional probability and state transition probability of nodes from the calibrated SOR-DBN quantify the causal relationship between cognitive factors, which is embedded into the cognitive GCN module to extract the characteristics of the influence mechanism of emotions on driving behavior. The CARLA-SUMO joint driving simulation platform was built to develop dangerous pre-crash scenarios. Methods of recreating traffic scenes were used to naturally induce abnormal emotions. The experiment collected data from 26 participants to verify the proposed model. Compared with the model that only considers physical motion features, the prediction accuracy of the proposed model is increased by 68.70%. Furthermore, considering the SOR-DBN reduces the prediction error of the trajectory by 15.93%. Compared with other advanced trajectory prediction models, the results of CPSOR-GCN also have lower errors. This model can be integrated into active safety systems to better adapt to the driver's emotions, which could effectively reduce false alarms.**

## I. Introduction

With the rapid growth of car ownership, the number of traffic accidents remains high [1]. To reduce traffic accidents, the automotive industry has been developing the active safety system [2]. However, current active safety systems suffer from frequent false alarms. In 2019, NHTSA received 129 complaints specifically regarding false alarms of Autonomous Emergency Braking (AEB) [3]. In addition, in the 2022 China Intelligent Driving Test, all tested vehicles were reported to have false alarms [4].

One of the important causes of the false alarm is that the system ignores the effects of abnormal emotions (anger, sadness, and so on) of a driver on the vehicle trajectory. Active safety systems rely on vehicle trajectory prediction to determine collision risk and execute alarms. Currently, most vehicle trajectory predictions assume that the driver is in a normal state [5-7]. However, when a driver is experiencing an abnormal emotion, the vehicle trajectory becomes hard to predict, which causes false alarms.

Abnormal emotions have significant effects on vehicle trajectory. For example, when a driver is experiencing fright, there will be more violent braking and operating errors, resulting in sudden changes in the trajectories [8, 9]. In addition, anger can significantly increase the fluctuation range of vehicle trajectories [10-12]. When a driver is angry, the vehicle trajectories show a more severe lateral deviation and a more violent longitudinal acceleration [10, 11]. Moreover, anger may lead to aggressive driving behavior [10, 13, 14]. An angry driver tends to drive more recklessly than normal, with a higher frequency of lane changes, overtaking, and violations. These effects of abnormal emotions make the vehicle trajectories hard to predict, particularly in a scenario with strong interactions.

Some researchers have tried to improve vehicle trajectory prediction by considering emotion. They regard emotional state as an important predictor variable [15, 16], and results show that the prediction accuracy has been improved. However, these model structures deviate from the driver's cognitive process, and ignore the causal relationship between emotion and other cognitive factors, resulting in limited prediction accuracy. It has been demonstrated that whether a specific emotion improves or degrades a specific decision depends on causal relationships among the cognitive and motivational mechanisms triggered by each emotion [17].

To overcome these limitations, this study aims to incorporate abnormal emotions into vehicle trajectory prediction in a manner consistent with the human cognitive process. CPSOR-GCN, a trajectory prediction model that integrates both cognitive and physical features, is proposed to improve prediction accuracy. Specifically, this research used Stimulus-Organism-Response (SOR) theory [18] to model cognitive process. SOR is an important theoretical reference for understanding how internal states, such as emotion, affect driving behavior under external stimuli. The main contributions of this study are as follows:

This research is funded by Shanghai Science and Technology Plan Project(23692123300). (Corresponding author: Lishengsha Yue.).

Lanyue Tang, Lishengsha Yue, Jian Sun, and Auhui Fu are with the Key Laboratory of Road and Traffic Engineering, Ministry of Education, Shanghai 201804, China(e-mail: 2133393@ tongji.edu.cn 2014yuelishengsa@tongji.edu.cn; sunjian@tongji.edu.cn; 2133397@ tongji.edu.cn).

Jinghui Yuan is is an R&D Associate Staff member in the Applied Research for Mobility Systems (ARMS) group at the Oak Ridge National Laboratory (ORNL), USA (e-mail: yuanj@ornl.gov).

(1) An online emotion induction method is adopted to obtain the vehicle trajectory when a driver experiences abnormal emotions. Unlike the offline induction method that interrupts simulated driving [19-21], the proposed method evokes the participants' abnormal emotions in traffic scenarios, which maintains the continuity and the authenticity of the experiment.

(2) CPSOR-GCN is proposed to predict vehicle trajectory considering the effects of emotion. Different from previous models that didn't consider a driver's cognitive process [16, 22-26], the proposed model constructs the effects of emotion based on SOR cognitive theory, and integrates both physical and cognitive features of the driver, which significantly improves the model accuracy.

The paper is organized as follows. Section 2 reviews the related works. Section 3 introduces the CPSOR-GCN model. Section 4 describes the data collection method and the driving simulator experiment. Section 5 presents the results. Section 6 discusses the model's limitations and concludes the study.

## II. Literature Review

### A. Deep-learning-based trajectory prediction model

Most deep-learning-based models make predictions based on physical motion features of the vehicle. In existing studies, CNN and the GCN are usually used for spatial feature extraction considering interaction factors and map information [25, 27-29]. RNNs and LSTMs are used to consider temporal features of vehicle physical trajectories[7, 30]. Among these four models, the application of the GCN model has achieved higher prediction accuracy in many practices, as it can better extract interactive features between vehicles [31]. Xin Li et al. [32] developed a GCN that used a graph based on the Euclidean distance to represent vehicles' interactions. Then, the extraction of interaction features was realized by graph convolution. Jiyao An et al. [22] further considered dynamic graphs in GCN by setting edge weights to continuous values related to the Euclidean distance. This approach better represents the strength of interaction between vehicles. In general, most of the deep learning models only consider the spatiotemporal characteristics of the physical motion of the vehicle. They ignore the significant effect of cognitive factors on trajectories, especially emotion[8, 10-12]. Consequently, because of missing important information regarding the causal relationship between emotion and other contributing factors, with these models it would be hard to explain some uncertainties in vehicle trajectories, even with complicated model structures.

### B. Probability-based trajectory prediction model

Some researchers consider the driver's emotional state in probability-based trajectory prediction models. They divide driving behavior into several fixed stages, and use transition probability between states to quantify the causal relationship between emotional states and other stages [31, 33]. In these works, probabilistic inference methods, such as Dynamic Bayesian Inference, are verified to be more favorable approaches for quantifying causal relationships. In one such study, Gill et al. [26] divided driver maneuvers into several phases. Through a probabilistic framework based on Dynamic Bayesian Inference, they couple the different phases and correlate them with the dynamic driving style. In another study, Danaf et al. [16] propose a probabilistic inference framework, which uses trait anger and state anger as latent variables, and the driver's decision-making and driving performance as observed variables. Nevertheless, the frameworks in these studies typically lack the support from a consensus-based theory regarding a driver's cognitive process, which results in ongoing controversies.

### C. Cognitive Theory describing the influence mechanism of emotion

The forementioned models lack convincing theory to describe drivers' cognitive processes, which limits the model accuracy. Driver behavior can be modeled using the Stimulus-Organism-Response (SOR) theory, which is a fundamental theory in behavioral psychology [18]. The SOR theory offers a crucial theoretical reference for comprehending how internal states, such as emotional state, impact driving behavior under external stimuli. This theory shares many similar features with phenomena observed in driving scenarios. Studies have found a significant link between traffic conditions and drivers' states corresponding with the process of stimulus to the organism [34]. Abnormal emotional states, such as anger or fright, are associated with aggressive driving behavior or mishandling [15]. These are summarized as the organism-response causal chain. The SOR theory has been widely used in the analysis of distracted driving behavior and the prediction of purchasing behavior [35].

## III. Methods

### A. Preliminaries

#### 1) Trajectory Prediction Problem

The problem of vehicle trajectory prediction is to use the physical motion and cognitive feature $\mathrm{F} = \{p_t^k, c_t^k \mid \mathrm{t} \in T_p, \mathrm{p} \in \mathrm{P}, \mathrm{c} \in \mathrm{C}, \mathrm{k} \in \mathrm{K}\}$ of the past time $\mathrm{t} \in T_p$ to predict the trajectories of the future $\mathrm{t} \in T_f$ period, $Y_{predict} = \{y_t^k \mid \mathrm{t} \in T_f, \mathrm{k} \in \mathrm{K}\}$. Here, $y_t^k = \{pos_{t_0}^k, pos_{t_0+1}^k, pos_{t_0+2}^k, \dots, pos_{t_0+T_f}^k\}$ represents the sequence of trajectory points for vehicle k within the prediction time period. Each $pos_t^k = (x_t^k, y_t^k)$ represents the spatial coordinates of the vehicle's position.

#### 2) Physical Features

Physical features related to a trajectory can be represented as $\mathrm{F} = \{p_t^k \mid \mathrm{t} \in T_p, \mathrm{p} \in \mathrm{P}, \mathrm{c} \in \mathrm{C}, \mathrm{k} \in \mathrm{K}\}$ . $T_p$ denotes the time intervals of the historical input. K represents the set of all vehicles, including the ego vehicle and the interacting vehicles. $p_t^k = \{x_t^k, y_t^k, v_t^k, a_t^k \mid \mathrm{t} \in T_p, \mathrm{k} \in \mathrm{K}\}$ captures the physical motion features of the vehicle, including the positional coordinates, velocity, and acceleration of the vehicles.

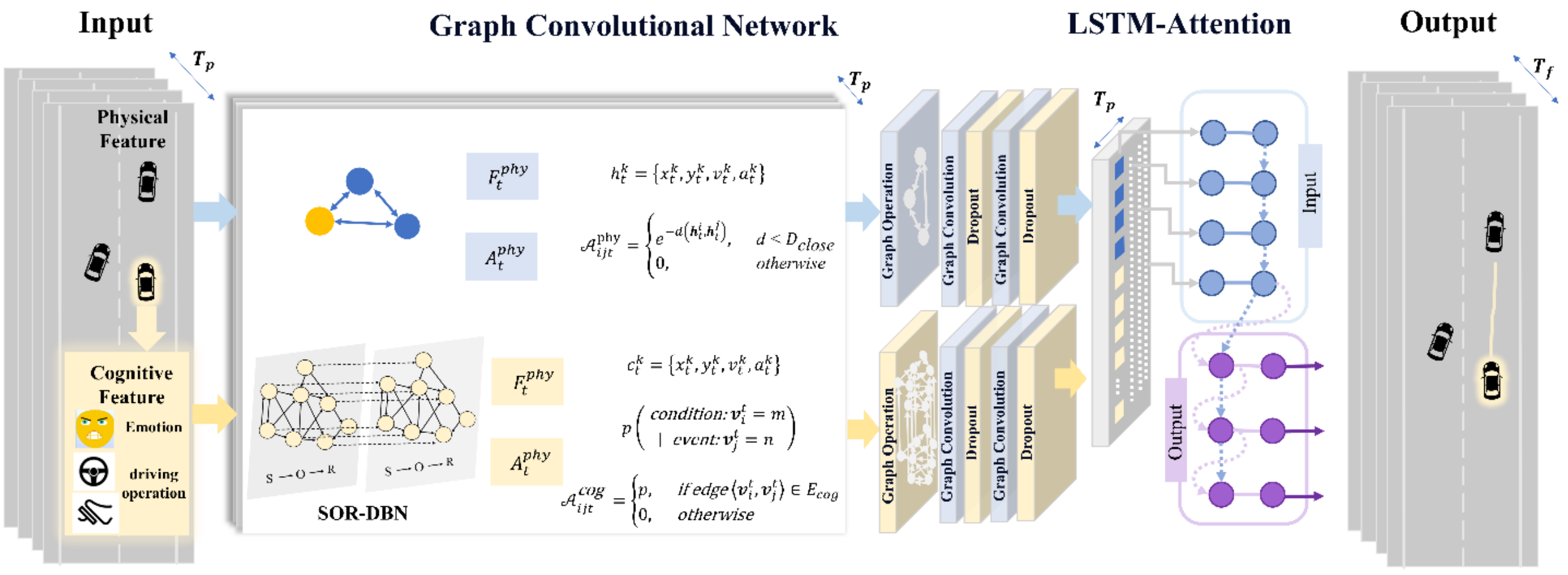

**Fig. 1.** The structure of the CPSOR-GCN model

*3) Cognition Features*

The broad cognitive process is generally considered to be the process in which an individual's internal state determines his or her behavior under the stimulus from the environment [35] . Therefore, this study defines driver cognitive factors related to trajectory to include three parts: stimulus from the environment, individual internal state, and the behavior as a response, which are specifically expressed as:

$$C_t = \{Npc_a_t,\ Risk_grade_t, Emo_cluster_t, Ego_a_t, Sub_style_t, Obj_style_t, Maneuver_t,\ Behavior_t | \mathrm{t} \in T_p\} \quad (1)$$

Stimulus includes the danger degree of the current environment and the actions of the interacting vehicle. $Risk_grade_t$ represents the degree of danger at time t by evaluating the conflict risk. The interactive vehicle's actions can be abstractly categorized as either approaching or avoiding behaviors [36]. Therefore, the acceleration of the interactive vehicle denoted as $Npc_a_t$ (the interactive vehicle acts as an NPC in a driving simulator experiments) is used to characterize it.

Individual internal state includes the driver's emotion, attitude, and subjective driving style. $Emo_cluster_t$ represents the driver's emotion at time t, including three states: anger, neutral, and fright in our study. $Ego_a_t$ represents the driver's attitude towards the traffic status at time t, which refers to the driver's evaluation of whether the speed should be changed from a safety perspective [35]. Therefore, the driver's acceleration is used to express the attitude. $Sub_style_t$ represents the driver's subjective driving style, which refers to the driver's subjective view of his or her driving style as aggressive, neutral, or conservative.

Behavioral responses include longitudinal maneuvers and lateral maneuvers, denoted as $Maneuver = \{Man_longi, Man_lateral\}$. Man_longi includes three states: acceleration, speed maintenance, and deceleration, while Man_lateral includes turning left, going straight, and turning right. $Obj_style$ refers to the objective driving style shown by actual driving behavior, which is divided into three types, recorded as $\mathrm{Obj_style} = \{\mathrm{hasty}, \mathrm{moderate}, \mathrm{gentle}\}$ . $Behavior$ is the coupling of Obj_style and Maneuver, which is a description of driving maneuvers with style, such as hasty braking and steering.

## B. *Trajectory prediction scheme*

The CPSOR-GCN model integrates both the physical motion features and cognitive features. As shown in Figure 1, CPSOR-GCN consists of three main components: the physical GCN module, the cognitive GCN module based on SOR, and the LSTM-attention module.

## C. *Physical GCN module*

The physical GCN module extracts vehicle interaction features at the physical level. Interactions often occur when two vehicles are in close proximity, similar to social interactions in a network [6]. Thus, a graph $G_{phy} = \{V_{phy}, E_{phy}\}$ is used to represent the interactions between vehicles. The node set $V$ represents all vehicles in the traffic scene. Each vehicle has different states at different time steps. The node set $V$ is denoted as follows:

$$V_{phy} = \{h_t^k \mid \mathrm{k} \in \mathrm{K}, \mathrm{t} \in T_p\} \quad (2)$$

The features of each node include positional information as well as motion features such as velocity and acceleration, denoted as:

$$h_t^k = \{x_t^k, y_t^k, v_t^k, a_t^k \mid \mathrm{t} \in T_p, \mathrm{k} \in \mathrm{K}\} \quad (3)$$

In the physical GCN, the presence of interactions between two vehicles is determined by their Euclidean distance. When the distance between two vehicles is close enough, interactions occur, and the edges representing their interactions should be connected. Therefore, the edge set $E$ is described as follows:

$$E_{phy} = \{\boldsymbol{h}_t^i \boldsymbol{h}_t^j \mid d(\boldsymbol{h}_t^i, \boldsymbol{h}_t^j) < D_{\mathrm{close}}\} \quad (4)$$

Where $d$ denotes the Euclidean distance function. For any given time step, the adjacency matrix in the physical layer graph input is defined as:

$$\mathcal{A}_{ijt}^{\mathrm{phy}} = \begin{cases} e^{-d(v_i^t, v_j^t)}, & \text{if edge } \langle \boldsymbol{v}_i^t, \boldsymbol{v}_j^t \rangle \in E_{phy} \\ 0, & \text{otherwise} \end{cases} \quad (5)$$

## D. Cognitive GCN module

### 1) SOR Framework

The Cognitive GCN module extracts the mechanism of driver emotion on driving behavior, which is required to reflect the driver's cognitive process. There are causal relationships between key factors in the driver's cognitive process, and the strength of each causal relationship is also evolving in real time. In this study, the nodes of the graph model in Cognitive GCN are cognitive elements from the SOR framework. SOR is a classic theory that describes the influence mechanism of environmental factors on human internal states (such as emotion) and subsequent behaviors [18]. Specifically, S (stimulus) refers to the influence of the environment in which the individual participates [35]. O (organism) generally refers to an individual's internal state that is affected by stimulus, often including the attitude and emotion [37]. R (response) refers to an individual's behavior response.

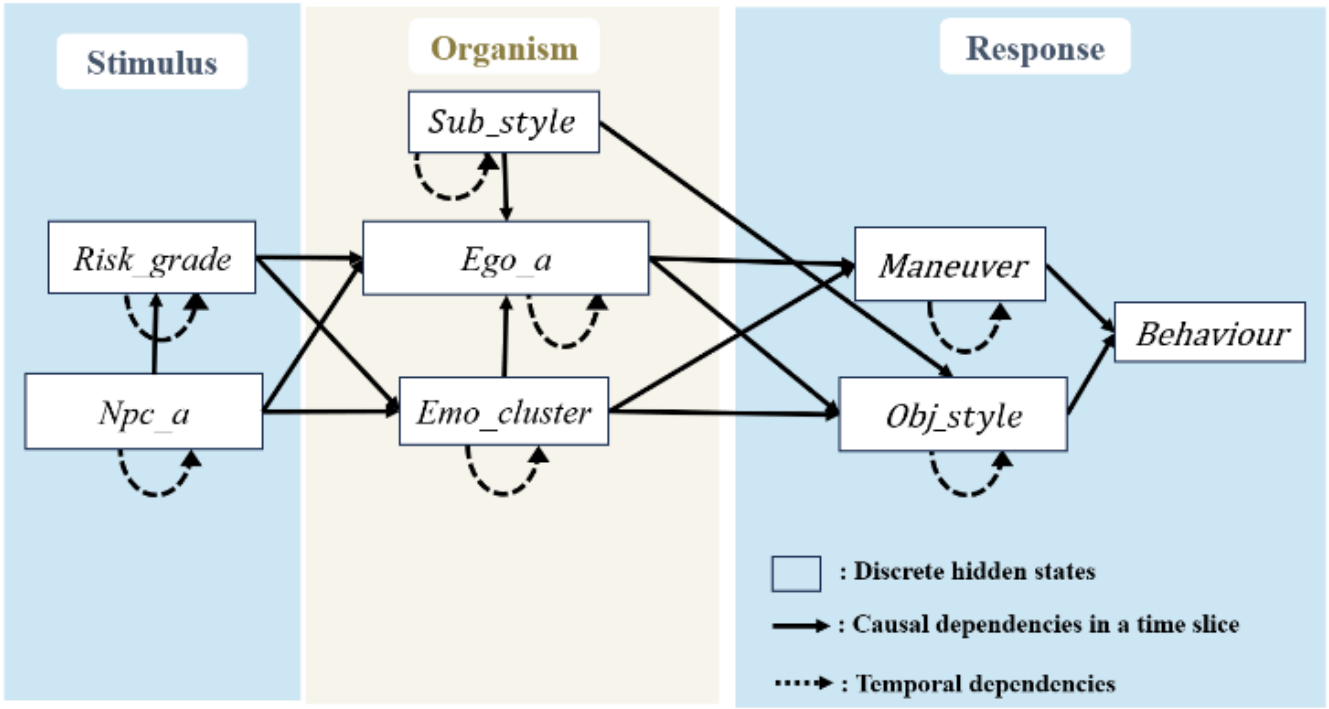

**Fig. 2.** The framework of SOR-DBN

### 2) Feature Tensor

The cognitive features in $C_t$ are used as nodes of graph data in the cognitive GCN model. In this section, the calculation method of node status will be introduced. In stimulus caused by the environment, each state of the $Risk_grade$ node is a risk level divided according to the time to collision (TTC) between vehicles [38]. TTC is estimated based on the current position, speed, and heading angle of the vehicles. In addition, the acceleration of the interactive vehicle is discretized at intervals of $0.2\mathrm{m/s^2}$ as the state of $Npc_a$.

In the individual internal state, the data recorded by the emotion scale are used to obtain the state value of the Emo_cluster through the kmeans clustering method. Sub_style is classified based on driving style scores obtained from the Driving Style Scale [39]. The higher the score, the more aggressive the driver's subjective driving style. The acceleration of the vehicle is discretized at $0.2\mathrm{m/s^2}$ as the state of $Ego_a$.

As a behavioral response, Man_Lateral in $\text{Maneuver} = \{\text{Man_longi}, \text{Man_Lateral}\}$ is divided into left turn, straight, or right turn according to the steering wheel angle at each time point with a threshold of ±4° [40]. Throttle force, brake force, and steering wheel angle are used to cluster the status of Man_longi and Obj_style. Then, the status of $\text{Behavior} = \{\text{Obj_style}, \text{Maneuver}\}$ node is obtained. The status division threshold standards and corresponding status of each node are shown in Table 1.

TABLE I
CALCULATION OF THE NODES' STATUSES

| Variable | Symbol | States |
|---|---|---|
| Risk grade | $Risk_grade$ | Safe (TTC[1] $> 2s$)<br>Moderate conflict ($1.5s \leq \text{TTC} \leq 2s$)<br>Danger ($\text{TTC} \leq 1.5s$) |
| The acceleration of interactive vehicle (Npc) | $Npc_a$ | Split into segments with an interval gap ($0.2\ \mathrm{m}/s^2$) |
| The acceleration of ego vehicle | $Ego_a$ | Split into segments with an interval gap ($0.2\ \mathrm{m}/s^2$) |
| Emotion | $Emo_cluster$ | Anger; neutral; fright |
| Subjective driving style | $Sub_style$ | Aggressive; neutral; conservative |
| Objective driving style | $Obj_style$ | Gentle; moderate; hasty |
| Longitudinal maneuver | $Man_longi$ | Accelerate<br>Speed maintenance<br>Decelerate |
| Transverse maneuver | $Man_Trans$ | Left turn (θ[2] $< -4°$)<br>Straight ($-4° \leq \theta \leq 4°$)<br>Right turn ($\theta > 4°$) |

[1] TTC: Time to collision; [2] θ: The steering wheel angle of the ego vehicle

The statuses of Man_longi, Emo_cluster, and Drive_style are obtained through K-means clustering. Maneuver data has a strong correlation in time, which means that maneuver data from multiple time slices constitute a behavior. To determine the length of the time window suitable for behavior clustering, the autocorrelation coefficient (AC) of each lagged data with the current data is calculated. The AC represents the correlation between the data of the current moment and the data of the lagged moment. Taking the time series data braking force $b_t$ as an example, the calculation formula of AC at lag order k is as follows:

$$AC(k) = \frac{Cov(b_t, b_{t-k})}{Var(b_t)} \tag{6}$$

$Cov(b_t, b_{t-k})$ represents the covariance between the brake force in time points t and t-k. $Var(b_t)$ represents the variance of brake force at time point t. The spacing between samples is iteratively increased to determine the optimal downsampling rate until the autocorrelation coefficient at lag 1 is significantly smaller. According to the AC analysis results, 20 time steps of data are selected as a clustering sample.

The average and standard deviation of the ego vehicle's brake force, throttle force, and steering wheel angle within 0.8 seconds are respectively used to cluster the ego vehicle's state of $Man_longi$ and $Obj_style$.

For the state of Emo_cluster, the emotion scale AffectButton [41] is based on the three-dimensional emotion ring model. Emotion is recorded as three-dimensional values of {Pleased,

Aroused, Dominant}. "Pleased" relates to the positiveness versus negativeness of emotion; "aroused" to the level of activation; and "dominant" to whether the environment is imposing influence over people or the inverse [42]. Kmeans clustering is used to cluster all recorded emotion data into three states: anger, neutral, and fright.

*3) Adjacency Tensor*

The causal relationship between cognitive nodes is used as the edge weight of the graph data in cognitive GCN, forming the adjacency matrix. A SOR-DBN framework is constructed, which represents the causal relationship between cognitive nodes based on SOR theory. The conditional probabilities between nodes and the state transition probabilities between time slices in the calibrated SOR-DBN model are used to quantify the causal relationship. The cognitive mechanism describing how emotions influence driving behavior can be described as the causal relationship between cognitive nodes. This cognitive mechanism satisfies the following Bayesian properties:

(1) The driver's cognitive mechanism remains basically constant at each time step (solid arrow in Figure 2).

(2) The states of cognitive nodes have a delayed effect over time. The current cognitive node state is related to the state of the previous time step (dashed arrow in Figure 2).

(3) There are causal relationships between cognitive nodes, whose strength changes over time.

The Bayesian network (BN) model $B_t$ (connected by solid lines) is shown in Figure 1, whose structure is a graph G = ($c_t$,E). The node set $c_t = \{c_t^1, c_t^2, \dots, c_t^l\}$ corresponds to the cognitive nodes in $C_t$ (formula). The edge set E represents whether there is a causal relationship between two nodes. The BN parameter $\theta'$ includes the conditional probability distribution between nodes, noted as $\theta' = P\left(c_t^i \mid Pa\left(cs_t^i\right)\right)$, where $Pa\left(c_t^i\right)$ is the parent node combination of node $c_t^i$. Taking node $Emo_cluster_t$ as an example, the corresponding parameter is represented by:

$$\theta_{tjm}^{\prime Emo_cluster} = \mathrm{P}(Emo_cluster_t = \mathrm{m} \mid (Risk_grade_t, Npc_a_t) = \mathrm{j}) \tag{7}$$

The parent nodes of node $Emo_cluster_t$ include $Risk_grade_t$ and $Npc_a_t$. J represents all possible value combinations of node $Risk_grade_t$ and node $Npc_a_t$. $\theta_{tjm}^{Emo_cluster}$ represents the conditional probability distribution of all states of node $Emo_cluster_t$ under all values of the parent node.

On the basis of $B_t$, the dynamic Bayesian network (DBN) considers the transfer network from $B_{t-1}$ to $B_t$, which is recorded as $B\left(\mathrm{B_t}, \vec{B}\right)$. $\vec{B}$ defines the state transition probability distribution between nodes in two adjacent time slices (connected by dotted lines). Correspondingly, in the $Emo_cluster_t$ node, the state transition probability is expressed as:

$$\vec{\theta}_{tjm}^{Emo_cluster} = P(Emo_cluster_t = m \mid Emo_cluster_{t-1} = j) \tag{8}$$

Where $j$ is all possible values of $Emo_cluster$ at time t-1. The state transition probabilities of nodes are used to quantify causality in time series. The delay effect of cognitive nodes can be well described. Combined with the conditional probability distribution $\theta_{tjm}^{\prime Emo_cluster}$ in each time slice, it can describe the changing trend of the driver's emotion under a specific stimulus (the specific value combination of $Risk_grade_t$ and $Npc_a_t$).

The parameter θ of DBN is the joint probability distribution obtained by multiplying all probabilities of nodes, recorded as:

$$\theta = \prod_{i=1}^{l} P\left(c_t^i \mid Pa\left(c_t^i\right)\right) \tag{9}$$

The calibration of DBN includes two parts: structure learning and parameter inference. Bayesian Information Criterion (BIC) scoring and a hill-climbing algorithm were used for the SOR structure learning [43]. Maximizing the BIC serves as the optimization objective, while the hill-climbing algorithm allows for faster traversal of all possible structures. The specific steps of structure learning include:

**a)** Using certain edges that are randomly generated to form the initial model structure based on the SOR cognitive framework as the expert's prior knowledge.

**b)** Learning the parameters of the current model using the maximum likelihood. The goal of parameter estimation is to find the parameter value $\theta^*$ that maximizes the probability of occurrence of the observed data. Given a set of data records $D = (\mathrm{D_1}, \mathrm{D_2}, \cdots, \mathrm{D_m})$ and parameter θ, the probability of data D appearing, denoted as P (D | θ), is called the likelihood of θ. Let θ vary in its domain, L (D | θ) = P (D | θ) is the likelihood function of θ. Then the estimation problem of DBN parameter θ is to find the value $\theta^*$ that makes L (D | θ) reach the maximum. $\theta^*$ is called the maximum likelihood estimate of θ.

$$\theta^* = \arg\max_{\theta} \mathrm{L}(\theta \mid \mathrm{D}) \tag{10}$$

**c)** Evaluating the complexity and fitness of the model structure by calculating the BIC scored and updating the optimal model structure. The formula for calculating the BIC score is as follows:

$$\mathrm{BIC} = \log(\mathrm{L}) - (0.5 * \mathrm{l} * \log(\mathrm{m})) \tag{11}$$

Where L represents the maximum likelihood function value, l represents the number of nodes in the model, and N represents the sample size of the dataset.

**d)** Repeating the above steps until all possible model structures have been explored.

For the optimal DBN network structure obtained, the probability distribution corresponding to the observation data is inferred through maximum likelihood estimation, as in step **(b).**

The transition probabilities of each node state are assigned as the edge weights between nodes in the cognition graph, denoted as:

$$\mathcal{A}_{ijt}^{\mathrm{cog}} = \begin{cases} p\left(\begin{array}{c}\text{condition: } \boldsymbol{v}_i^t = \mathrm{q} \\ \mid \text{event}: \boldsymbol{v}_j^t = \mathrm{j}\end{array}\right), & \text{if edge } \left\langle \boldsymbol{v}_i^t, \boldsymbol{v}_j^t \right\rangle \in E_{cog} \\ 0, & \text{otherwise} \end{cases} \tag{12}$$

Where,

$$p\big(\text{ condition: } \boldsymbol{v}_i^t = q \mid \text{ event: } \boldsymbol{v}_j^t = \mathrm{j}\big) \tag{13}$$

represents the probability of node $\boldsymbol{v}_i^t$ having a feature value of $q$ when node $\boldsymbol{v}_j^t$ has a feature value of $j$.

*4) Feature Operation*

GCN layers perform convolutional operations using the adjacency matrix and node features. The formula is as follows:

$$H^{(l+1)} = \sigma\left(\tilde{D}^{-1/2}\tilde{A}\tilde{D}^{-1/2}H^{(l)}W^{(l)}\right) \tag{14}$$

First, the adjacency matrix is normalized. Then, the node feature matrix of the previous layer, $H^{(l)}$, is multiplied with the normalized adjacency matrix, $\tilde{A}$, to obtain the propagated features. Prior to feature propagation, $H^{(l)}$ can be multiplied by the weight matrix, $W^{(l)}$, to introduce learnable parameters. Finally, an activation function $\sigma$ is applied to obtain the node feature matrix for the $l+1$ layer.

## *E. LSTM-attention module*

In the temporal dimension, the LSTM module captures the driver's temporal sequence features based on the output of the dual-layer GCN, as shown in Figure 3. The combination of the GCNs and the LSTM components allows for more effective extraction of the spatiotemporal features of vehicle interactions [44]. With the LSTM module, the temporal characteristics of the driver's cognitive mechanism and physical motion features can be better learned [45]. The Attention module allows the model to selectively focus on different parts of the historical data when predicting trajectories [46, 47]. The purpose of this layer is to highlight the key features by assigning learned weights to input vectors at different time steps. This allows predictive models to dynamically adjust attention and give more weight to data at important moments. Therefore, the prediction model can better adapt to changing trajectory features and cognitive mechanisms.

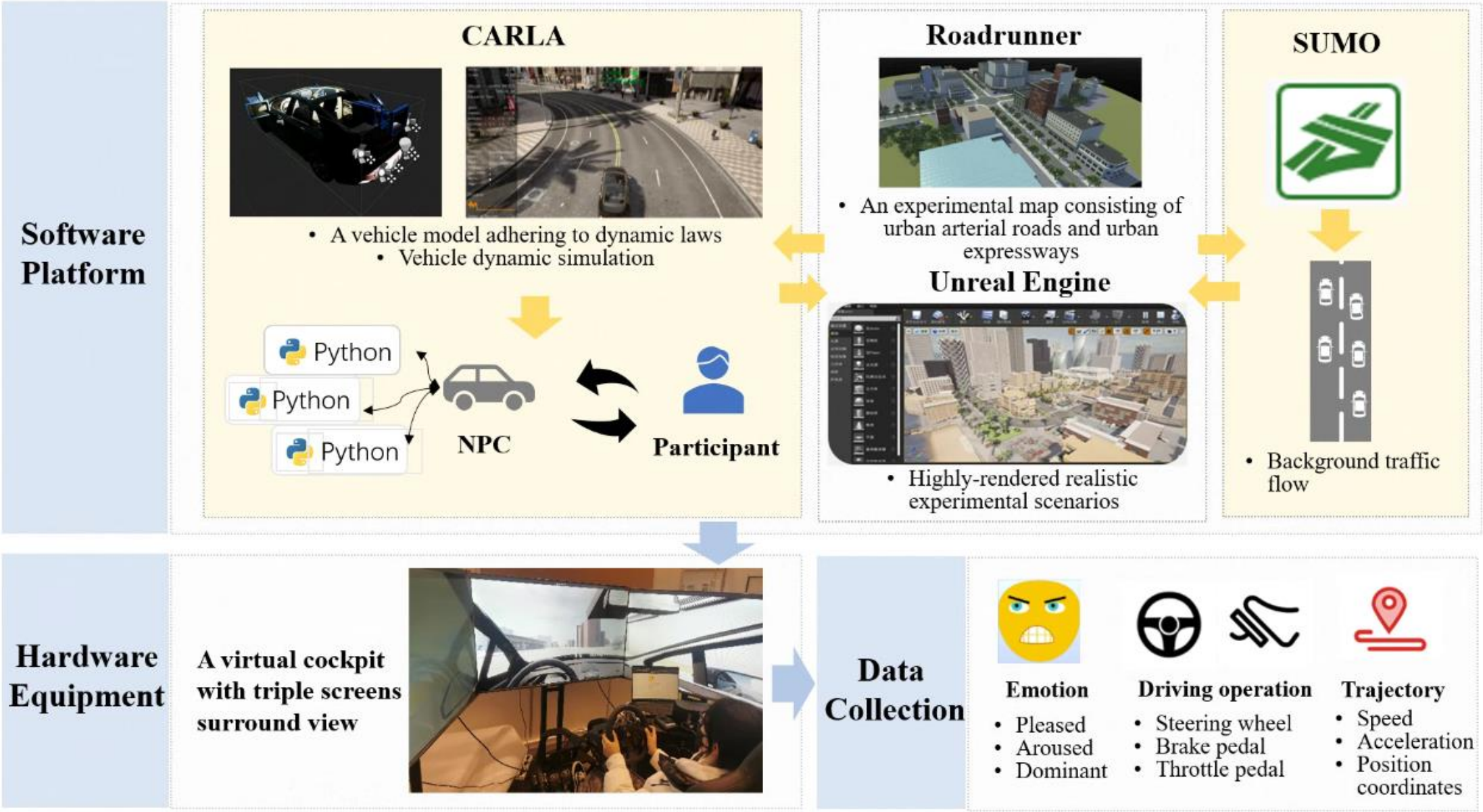

**Fig. 3.** The CARLA-SUMO co-simulation platform

TABLE II
SCENARIO SETTINGS

| Scenario ID | Scenario configuration | Scenario description |
|---|---|---|
| Scenario 1 (the sudden braking of the front car) | 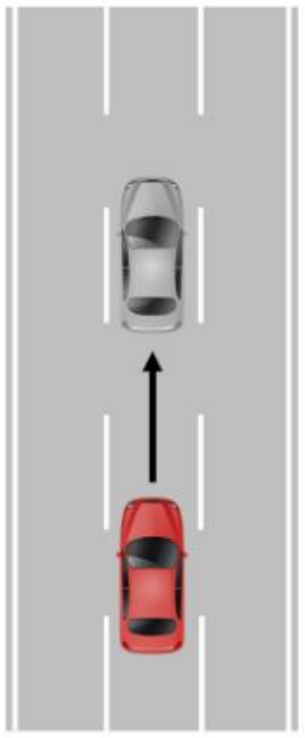 | At the beginning, the speed of the NPC in front is controlled through the program, so that the ego vehicle (driven by the participant) can follow it stably. When the participant reaches the targeted position, the NPC triggers the sudden braking. |
| Scenario 2 (vehicle cutting in) | 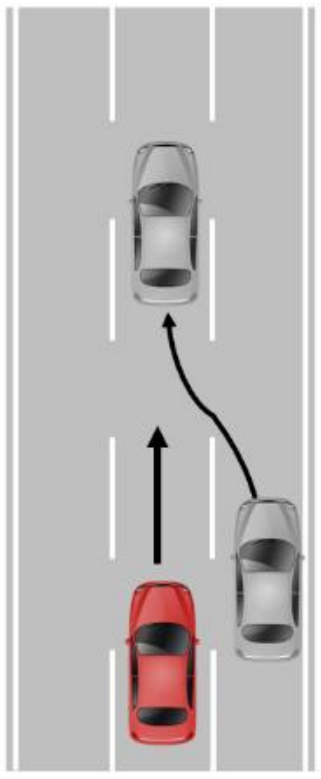 | NPC executes a lane-changing maneuver starting from an adjacent lane and ending in the ego vehicle's lane. A third-order Bezier curve [48] fits the cut-in trajectory. |
| Scenario 3 (a cyclist suddenly crossing the road) | 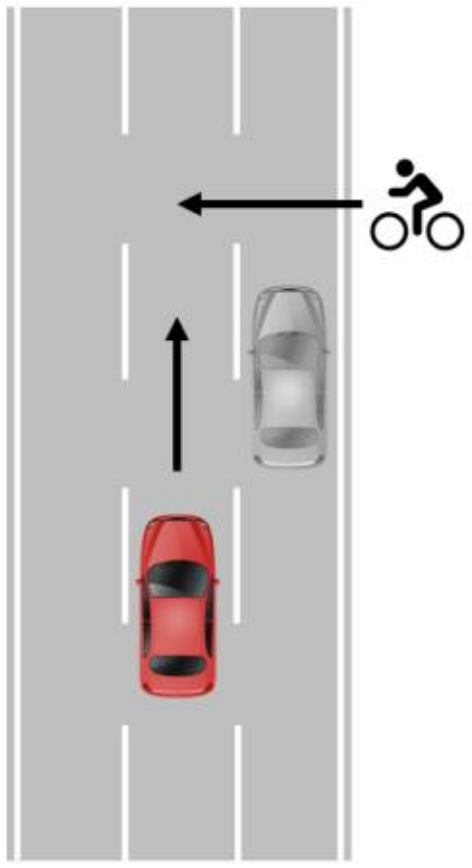 | A cyclist hides behind a car parked at the side of the road at the beginning. When the ego car passes by, the cyclist suddenly starts to cross the road. |
| Scenario 4 (unprotected vehicle left-turn) | 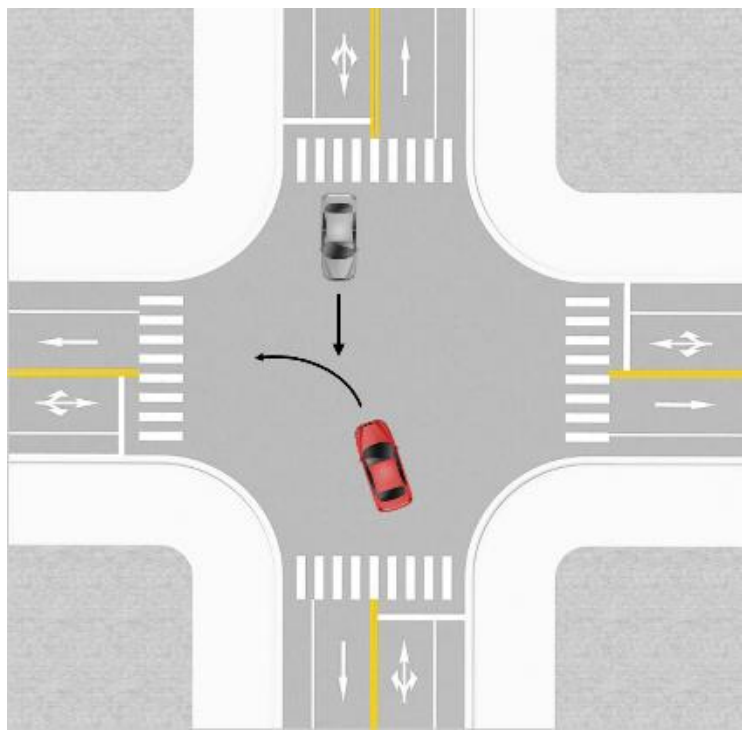 | The unprotected left-turn scenario is designed at an intersection without traffic lights. The ego car (red) will turn left and interact with a straight-going NPC. |

## IV. Data Collection

This study conducted a driving simulator experiment to collect vehicle trajectory data when a driver experiences abnormal emotions. An online emotion induction method was proposed to induce a driver's anger and fright. Data in pre-crash scenarios were collected.

### *A. Apparatus*

This study developed a CARLA-SUMO co-simulation platform to conduct the experiment, which is shown in Figure 3. CARLA provides vehicle dynamic models, and it allows participants to drive in a simulated environment. SUMO provides background traffic flow. A 3D map consisting of urban arterial roads and urban expressways was constructed by Roadrunner. A desktop computer with RTX 3080 graphics

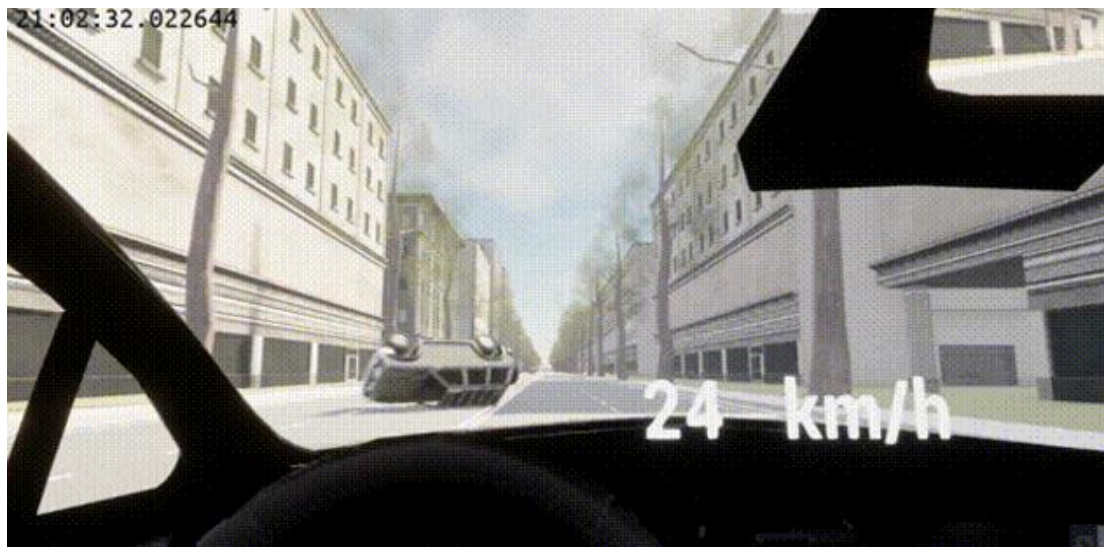

(a) Fright-Induction: Witnessing a traffic accident

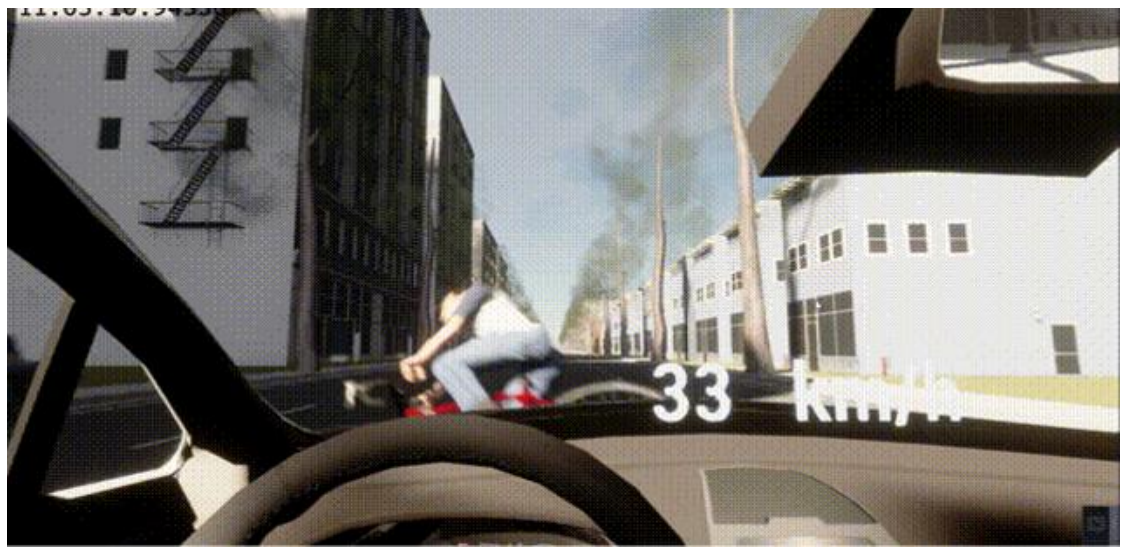

(b) Fright-Induction: Colliding with a pedal cyclist

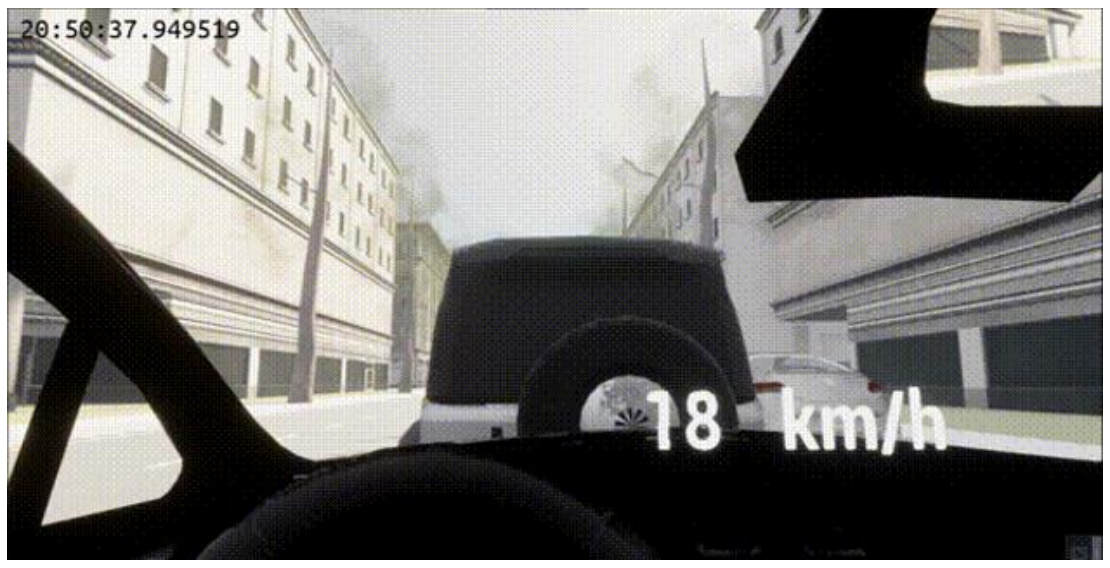

(c) Anger-Induction: Congestion

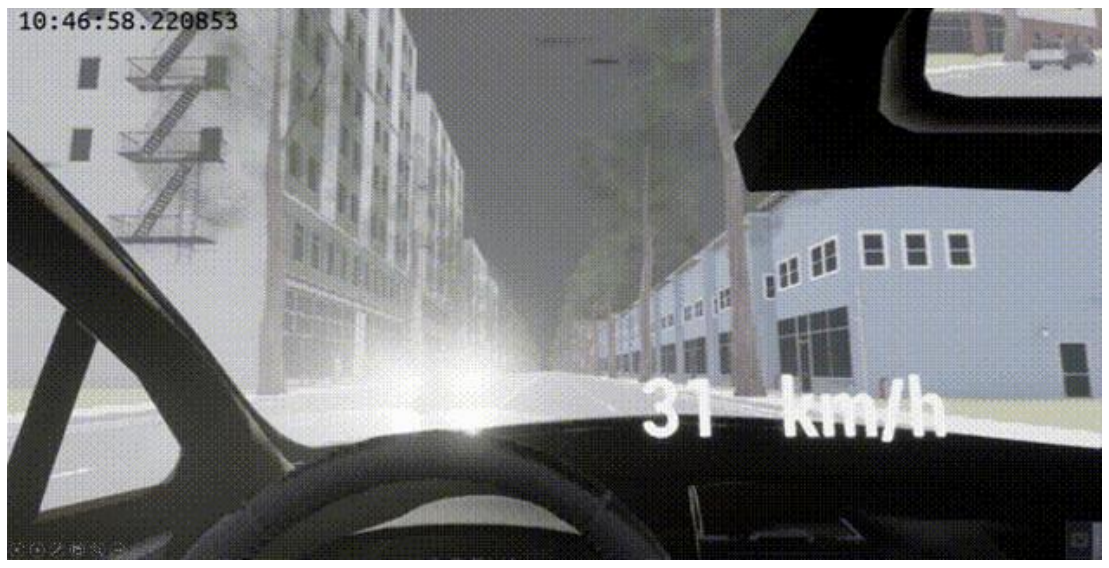

(d) Anger-Induction: The provocation of high beam lights of oncoming vehicles

**Fig. 4.** Experimental scenarios for emotion induction

cards was used to realize the simulated environment. The Logitech G923 steering wheel pedals were used for the driving control.

*B. Scenario design*

This experiment is based on the typical pre-crash scenarios reported by the NHTSA [49]. Specifically, the experimental scenarios include the sudden braking of the front car, the vehicle cutting in, the scenario of a cyclist suddenly crossing the road, and the vehicle executing an unprotected left-turn. Table 2 shows the design details of each scenario.

*C. Emotion induction*

An online emotion induction method was used to make a driver experience abnormal emotions of anger and fright. The traditional emotion induction method requires a participant to either watch videos or recall memories before the experiments. This offline method degrades the continuity of the experiment and the participant's immersive experience, which may lead to a lack of authentic vehicle trajectory. Instead, this study uses traffic scenarios to induce abnormal emotion. Based on previous investigation reports [15, 50], several traffic scenarios that often evoke anger and fright were replicated in the experiment (Figure 4).

Scenarios of congested traffic flow and the provocative behavior of flashing lights from an oncoming vehicle were replicated to induce anger. Wang et al. [51] found that more than half of road rage incidents were caused by the illegal use of high beams on rural roads or highways, and most of the accidents were caused by the high beams of oncoming cars. Chen et al. [50] analyzed the causal chain of drivers' road rage behavior by analyzing the video data of driving in Shanghai. The results showed that the prevalence rate of congested driving was as high as 64.72% among the triggers of angry driving. Scenarios of witnessing accidents and crashes involving pedal cyclists were replicated to induce fright [52]. Scenarios of smooth and safe driving sections were used for neutral emotion so that the participants could calm themselves down.

Before and after the pre-crash scenario and the emotion induction scenario, a self-report scale called AffectButton [41] was filled out by participants to measure subjective emotion. AffectButton is in the form of a picture. Participants simply use the mouse to move and select an expression that matches their current emotion, and then their emotional states were recorded as three-dimensional values of {pleased, aroused, dominant}. AffectButton will not take up too much attention of the participant and can be completed within 1-2 seconds. Since the development of the AffectButton Emotional Self-Report Scale, it has been successfully applied in many research studies [53, 54].

*C. Participants and procedure*

Twenty-six participants (mean age = 22.7, SD = 1.64) were recruited into the experiment. All participants were required to have more than two years of driving experience (M = 2.7 years, SD = 0.82).

Before the experiment, participants were told to drive as they normally would and that they could quit the experiment at any time if they wanted. Meanwhile, each participant had a practice drive to be familiar with the simulator.

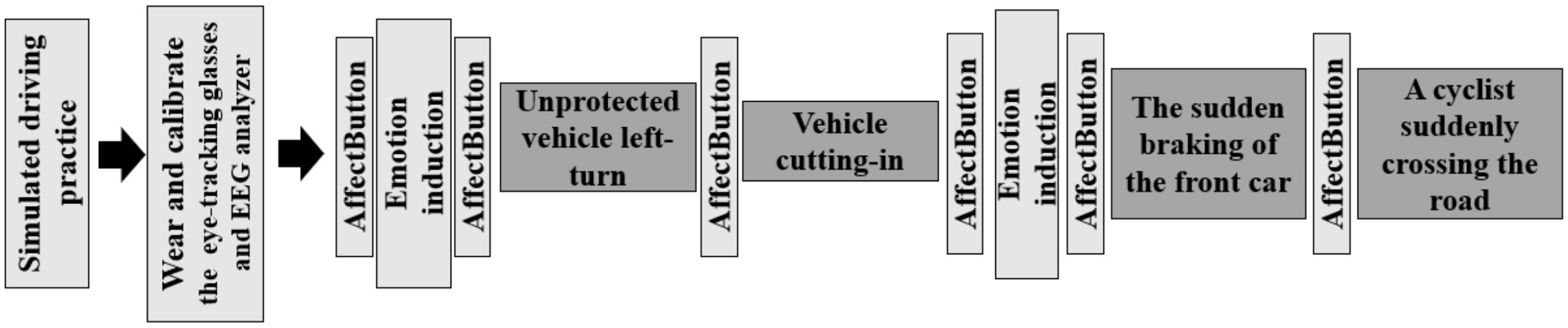

**Fig. 5.** The process of the experiment

During the experiment, participants were asked to drive in four experimental scenarios under each of the three emotions. In order to eliminate the carryover effect, the order of scenarios was counterbalanced with the Latin Sequence [55]. Each experiment lasted around 20 minutes, with enough rest time in between to avoid driving fatigue. The experiment process is shown in Figure 5.

## V. Results

### A. Effectiveness of emotion induction

We performed repeated-measures ANOVA on the emotion values collected before the induction and after the induction. Figure 6 shows that there are significant differences of emotion between the two time points. The {pleased,aroused,dominant} X = {-1,1,1} corresponds to extremely angry emotion, while the {pleased,aroused,dominant} = {-1,1,-1} corresponds to extremely frightened emotion. From the above results, it can be concluded that after being induced, the driver's emotion is significantly changed. This shows that emotion induction was effective.

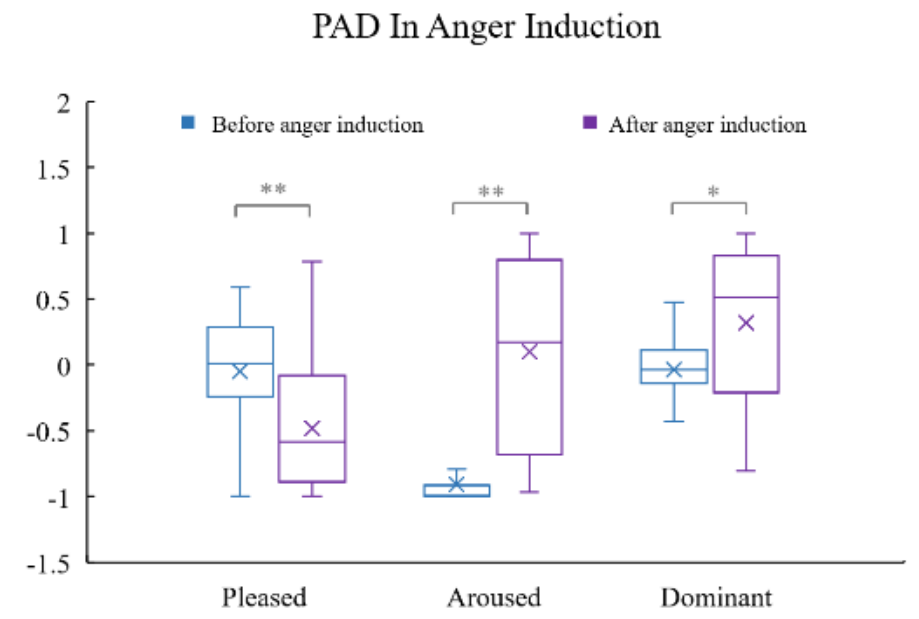

(a) Anger Induction

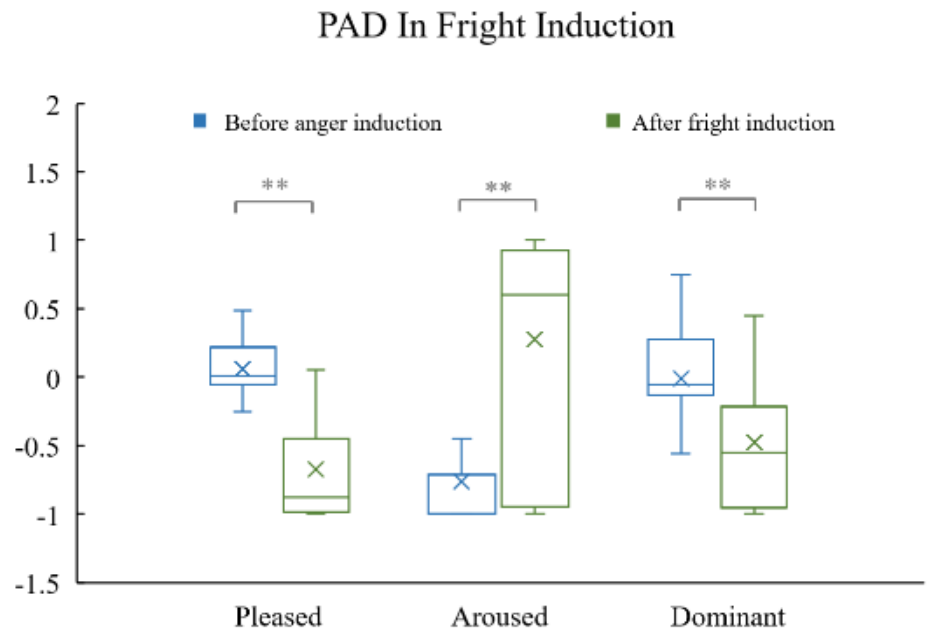

(b) Fright Induction

* $p$<0.05 ** $p$<0.01

**Fig. 6.** Significance test of emotion induction

### B. Model performance

In this section, the effectiveness of SOR theory as expert prior knowledge in extracting the mechanism of emotion on driving behavior is verified. Then in next section, ablation experiments are conducted to evaluate the effectiveness of the cognitive module based on the SOR-DBN structure in terms of trajectory prediction accuracy. Finally, the prediction accuracy of the proposed model is compared with other state-of-the-art (SOTA) models.

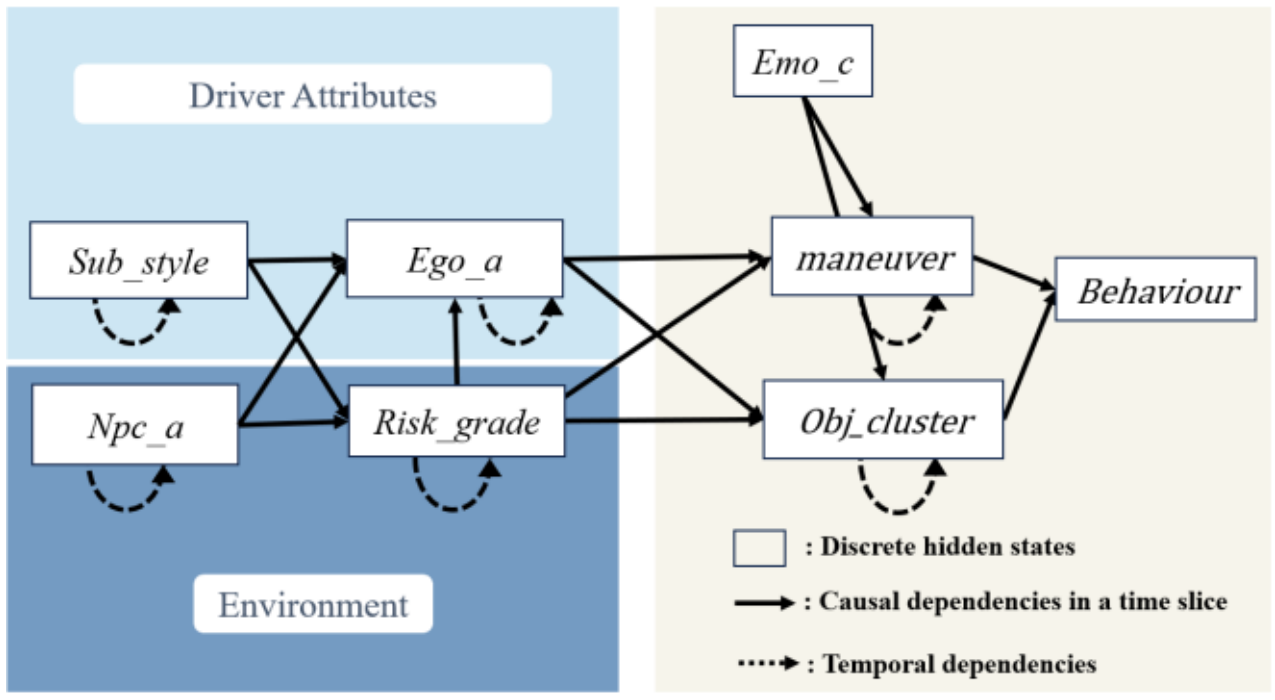

**Fig. 7.** The ordinary DBN

Accurately extracting the mechanism of emotion is crucial for predicting trajectories. The DBN structure lacking the prior knowledge of SOR experts is shown in Figure 7. The Ordinary DBN framework is constructed based on the discussion on the relationship between nodes in previous studies. These researchers believed that the driver's subjective driving style is an important characteristic of the driver, which is often placed at the position of the parent node as an important input [56]. The actions of the interactive vehicle are considered environmental factors as another parent node. When it comes to the influence of emotion on driving behavior, emotion is usually used as an exogenous variable, directly affecting the driver's final actions [16].

The two DBN structures with and without SOR as expert prior knowledge are compared in four scenarios. The final BIC score results are shown in Figure 8. The BIC score takes into account both the degree of fit and the penalty for structural complexity. Smaller BIC values indicate better structures, which utilize less information and yield higher fitting accuracy [57]. As can be seen from the Figure 8, the BIC scores of the SOR-DBN structure in the four scenarios are significantly smaller than those of the ordinary DBN structure, indicating that its degree of fit is still better when considering the penalty of structural complexity.

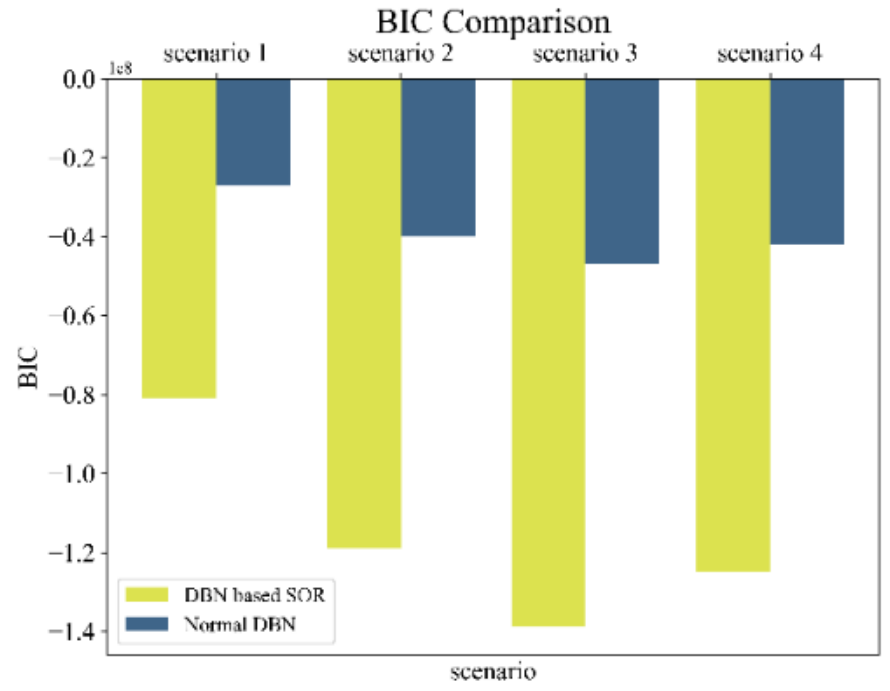

**Fig. 8.** BIC results of the ordinary DBN and SOR-DBN

Taking scenario 1 (the sudden braking of the front car) as an example, the conditional probability distribution P(Ego_a=p|Npc_a=q) of Ego_a=p occurring when Npc_a=q occurs, inferred from the two DBN frameworks, is shown in Figure 9.

As can be seen from Figure 9, the inference results of the SOR-DBN framework are closer to the distribution of real data. When considering the influence of the acceleration of the interactive vehicle (NPC) on the ego vehicle's acceleration ($P(Ego_a|Npc_a)$), the difference between the SOR-DBN and the ordinary DBN is that SOR-DBN describes the mechanism from environment to emotion and emotion to driving behavior, rather than directly treating emotion as an exogenous independent variable affecting behavior. The inference results verify the effectiveness of using the cognitive structure of SOR to consider the mechanism of emotion on the vehicle's acceleration. According to the chain rule of conditional probability, the parameter estimation process of the mechanism of $Npc_a$ affecting $Emo_cluster$ and then affecting $Ego_a$ is as follows:

$$P(Ego_a|Npc_a) = P(Npc_a) * P(Emo_cluster|Npc_a) * P(Ego_a|Emo_cluster) \quad (15)$$

Different from directly estimating $P(Ego_a|Npc_a)$, $P(Emo_cluster|Npc_a)$, $P(Ego_a|Emo_cluster)$ can be estimated to capture the causal relationship between $Ego_a$ and $Npc_a$ more accurately. SOR-DBN structure can allow a more detailed description of how the $Npc_a$ state leads to the occurrence of the $Ego_a$ state.

*C. Comparison of the trajectory prediction accuracy*

*1) Metrics*

In this section, we selected several evaluation metrics commonly used in trajectory prediction problems to assess the accuracy of the model predictions:

a) Root Mean Square Error (RMSE) is a measure of the square root of the average squared difference between predicted values and true values.

$$RMSE= \sqrt{\frac{1}{n}\sum_{i=1}^{n} \left(P_{pred}^{t}[i] - P_{true}^{t}[i]\right)^2} \quad (16)$$

b) Mean Absolute Error (MAE) is the average of the absolute difference between predicted values and true values.

$$MAE = \frac{1}{n}\sum_{i=1}^{n} \left|P_{pred}^{t}[i] - P_{true}^{t}[i]\right| \quad (17)$$

c) Average Displacement Error (ADE) represents the average L2 distance between the true values and predicted values for all prediction time steps.

$$ADE = \frac{1}{T}\sum_{t=1}^{T} RMSE^{t} \quad (18)$$

d) Final Displacement Error (FDE) measures the distance between the predicted final position and the true final position at the end of the prediction time horizon.

$$FDE = RMSE^{T} \quad (19)$$

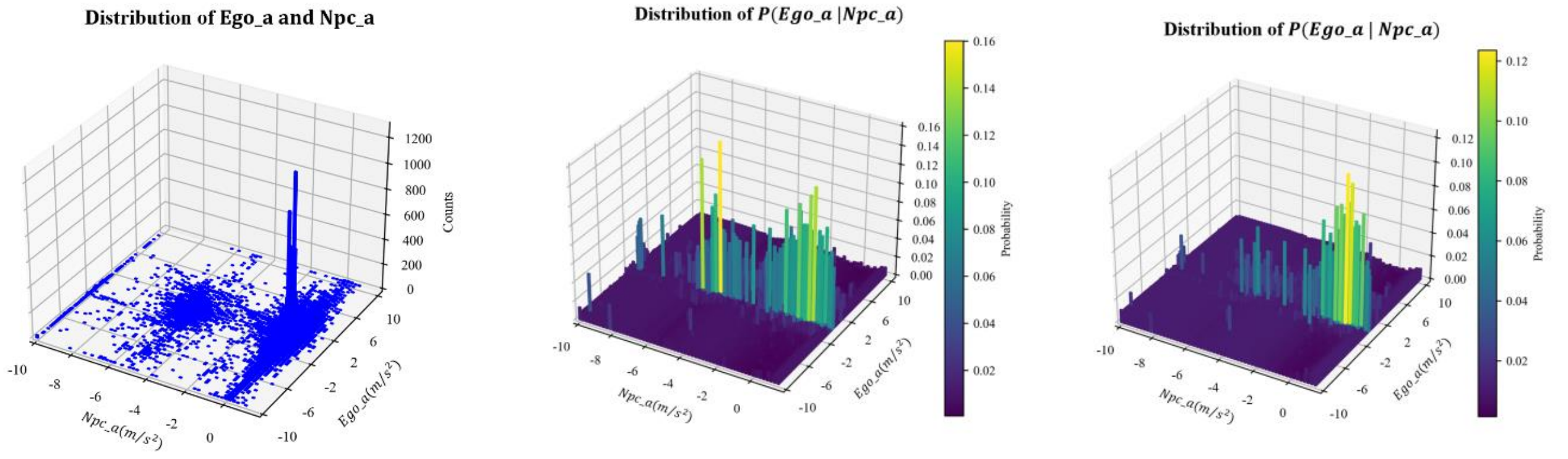

(a) Real counts

(b) Probability distribution inferenced by ordinary DBN

(c) Probability distribution inferenced by the SOR-DBN

**Fig. 9.** The distribution inferenced by ordinary DBN and SOR-DBN

2) Influence of SOR-DBN framework

In order to verify the effectiveness of the SOR-DBN-based cognitive GCN module, the following ablation experiments are designed, including three conditions: (a) the physical GCN prediction model (P-GCN) containing only the physical GCN module; (b) the cognitive-physical GCN prediction model (CP-GCN) based on the ordinary DBN without the prior knowledge of SOR; and (c) the cognitive-physical GCN prediction model (CPSOR-GCN) based on SOR-DBN. Under these three experimental conditions, the historical data of the past three seconds are used for prediction. The prediction time window is set to 1s, 2s, and 3s for comprehensive comparison. The indicators of RMSE, MAE, ADE, and FDE in the four scenarios are shown in Figure 10.

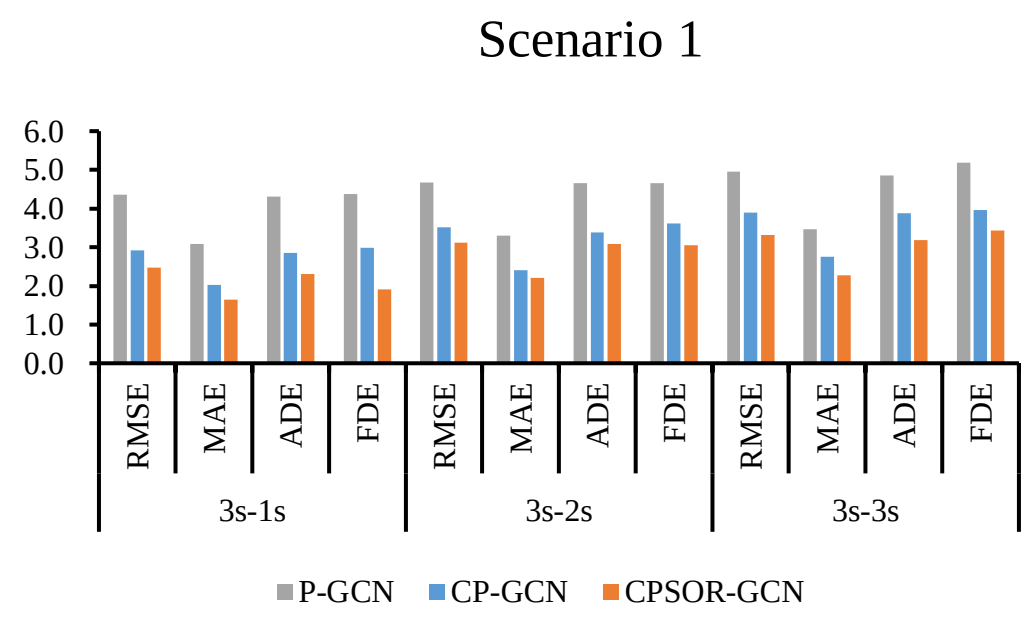

(a) Scenario 1: the sudden braking of the front car

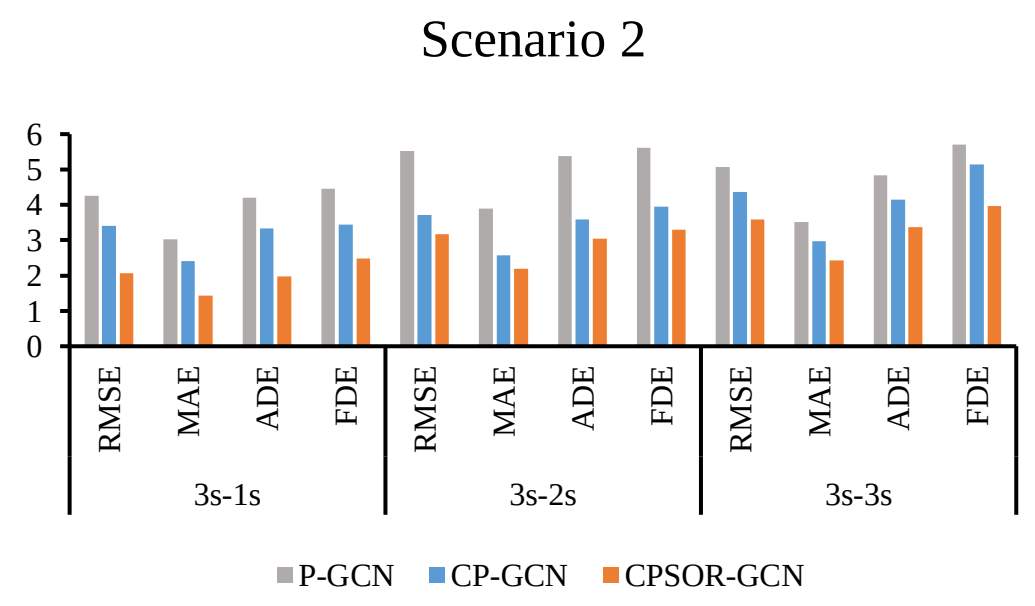

(b) Scenario 2: vehicle cutting in

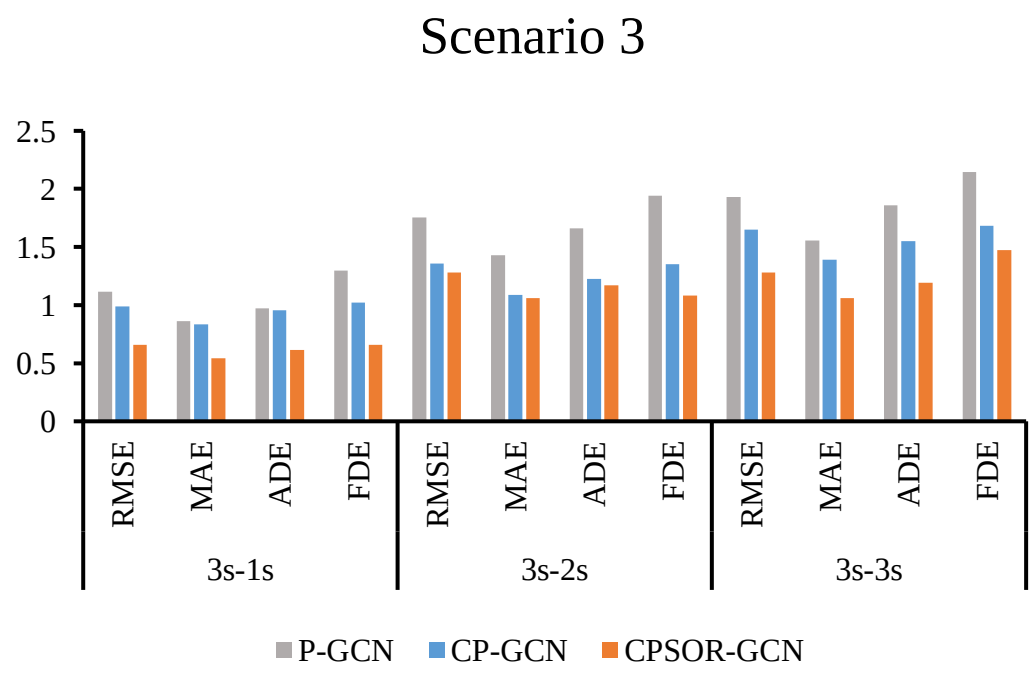

(c) Scenario 3: a cyclist suddenly crossing the road

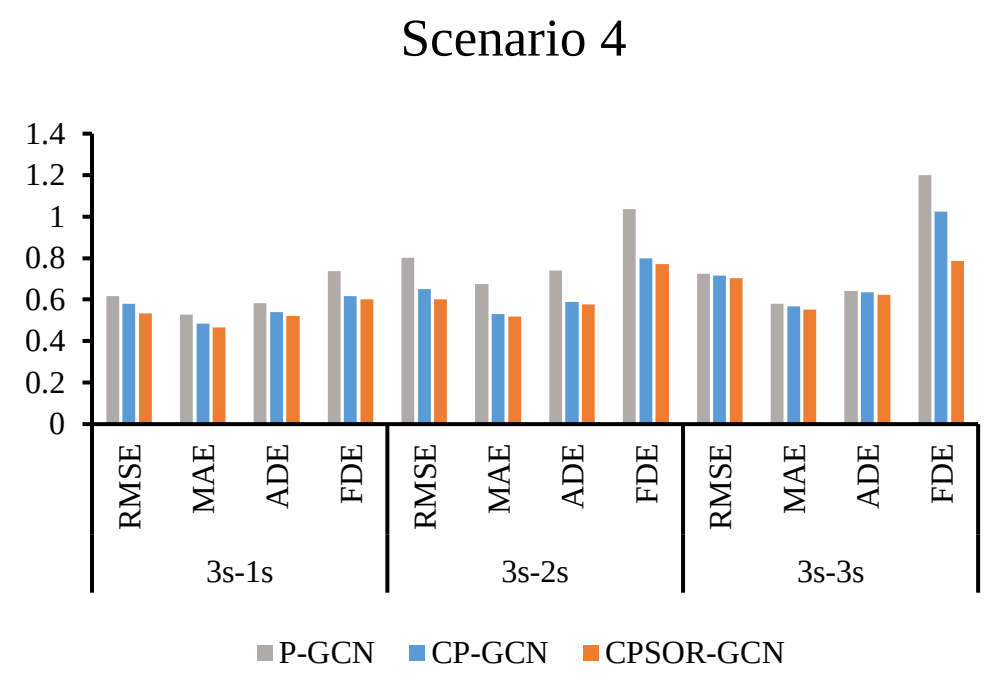

(d) Scenario 4: unprotected vehicle left-turn

**Fig. 10.** The results of ablation experiment

It can be seen from the experimental results that P-GCN, which is only based on the physical motion features of the vehicle, has the worst trajectory prediction accuracy. After adding cognitive features, the trajectory prediction error RMSE of the CP-GCN decreased by 26.22%, 16.22%, 33.41%, and 8.7% respectively, in the scenarios of sudden braking of the front car, vehicle cutting in, a cyclist suddenly crossing the road, and unprotected vehicle left-turn. The model prediction accuracy is significantly improved. As the prediction time window increases, the prediction error of the P-GCN, which only considers physical motion features, further increases. CP-GCN, which considers cognitive features of the driver, still maintains lower prediction errors in long-term predictions. Experimental results show that the CP-GCN model significantly reduces prediction errors by considering the driver's cognitive features.

It also can be seen from the results that the prediction error of the CPSOR-GCN model is significantly lower than that of CP-GCN. The ordinary DBN of CP-GCN does not consider the prior knowledge of SOR experts when extracting cognitive features. In the four scenarios, considering the SOR-DBN structure further reduced the prediction error RMSE by 13.55%, 23.77%, 20.48%, and 5.9%. As the prediction time window increases, the advantage of the CPSOR-GCN in prediction accuracy still remains. In addition, the CPSOR-GCN model has significantly lower FDE, which indicates that CPSOR-GCN can more accurately predict the end point of each sample trajectory. The effectiveness of the SOR cognitive framework is further verified by the significant advantage in trajectory prediction accuracy.

Figure 11 shows the predicted trajectories of CP-GCN and CPSOR-GCN when the No. 24 driver experienced angry, neutral, and frightened emotion, both under the condition that the 3s historical trajectory is used to predict the future 1s trajectory. The red dot in the picture marks the position of the ego vehicle when the front vehicle brakes suddenly.

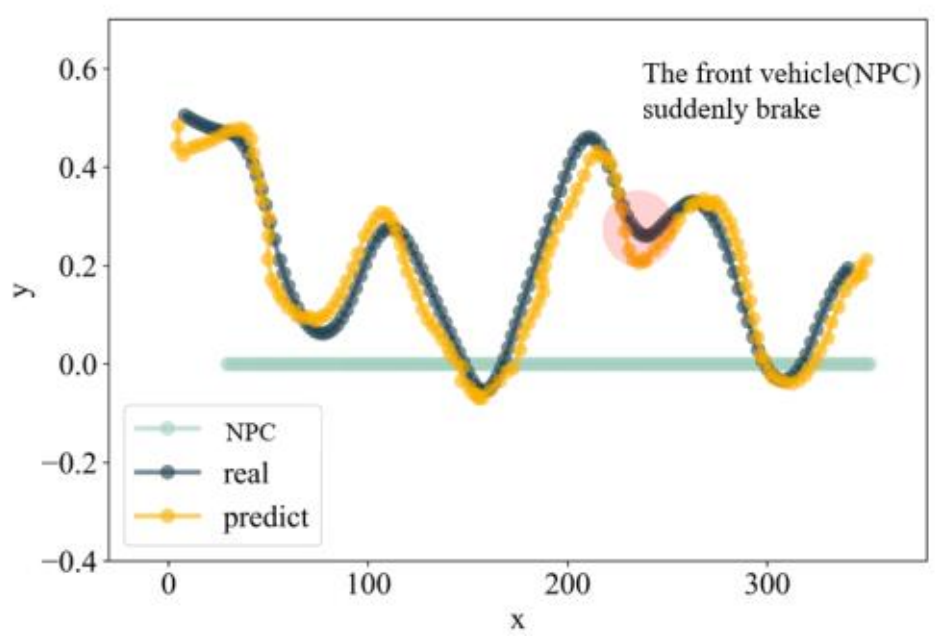

(a) CP-GCN: Neutral emotion

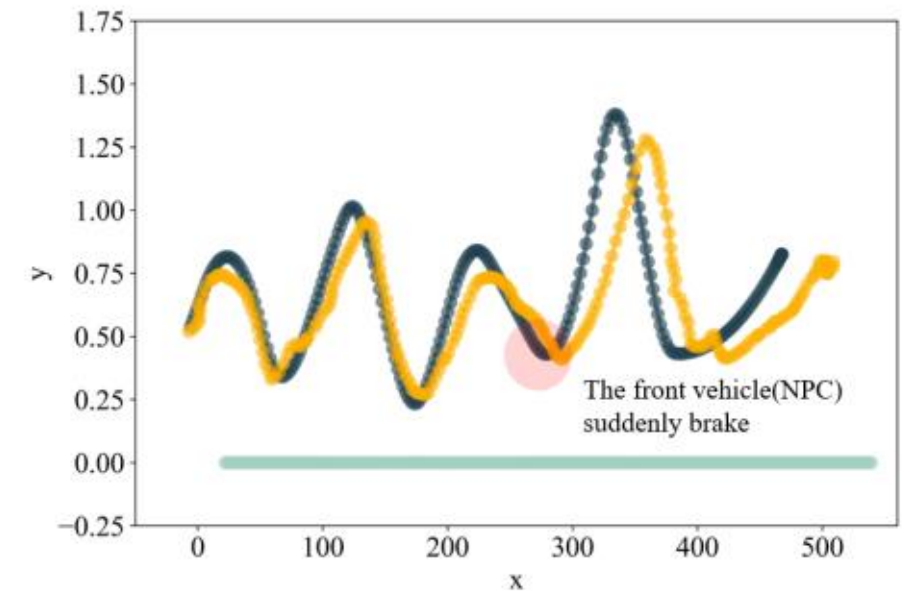

(b) CP-GCN: The emotion of fright

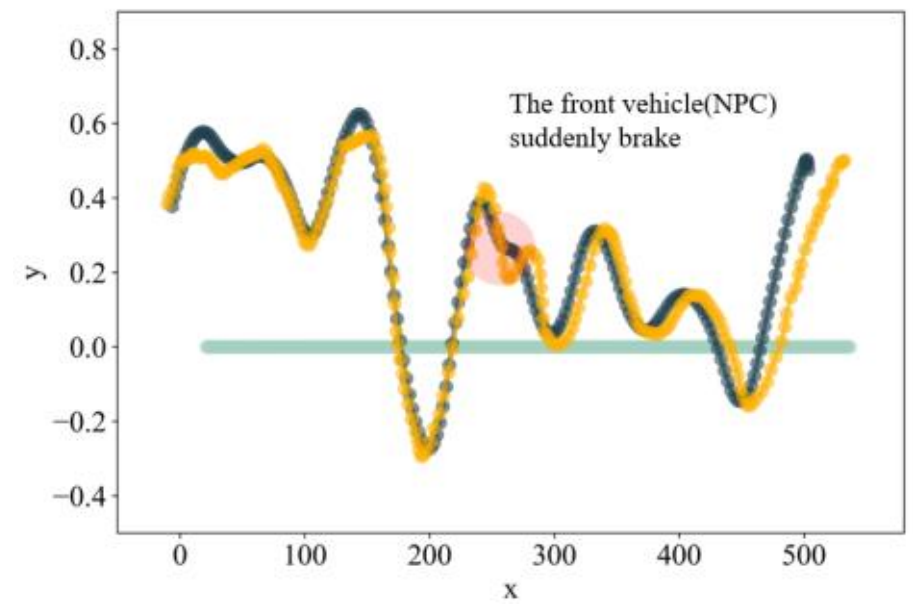

(c) CP-GCN: The emotion of anger

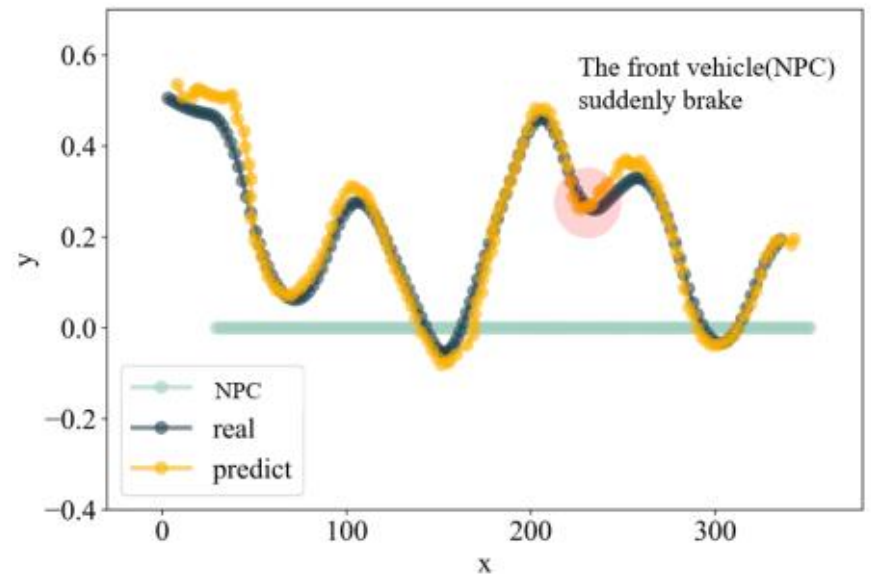

(d) CPSOR-GCN: Neutral emotion

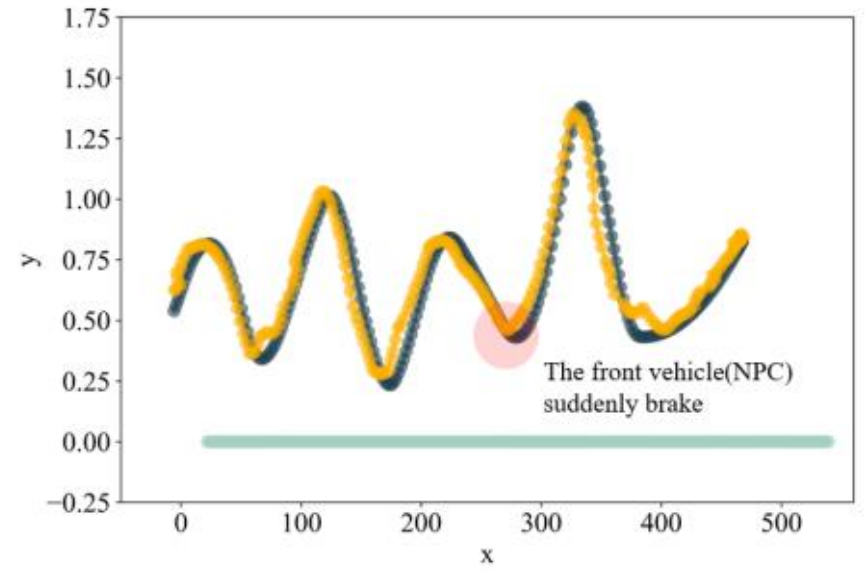

(e) CPSOR-GCN: The emotion of fright

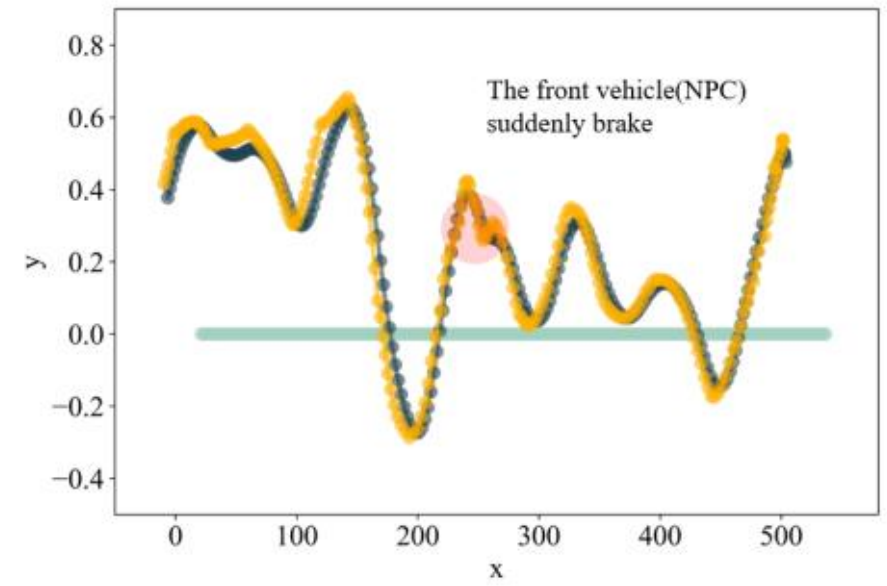

(f) CPSOR-GCN: The emotion of anger

**Fig. 11.** The results of predicted trajectory

It can be seen from the predicted trajectories that the CPSOR-GCN model has higher accuracy in predicting the avoidance behavior under the neutral emotion after the sudden braking event of the front vehicle. When the driver is in abnormal emotions of fright and anger, the prediction accuracy of the CPSOR-GCN model is significantly higher than that of the CP-GCN model. Compared with the ordinary DBN, the SOR-DBN cognitive framework has the following two advantages:

a) The SOR-DBN cognitive framework considers the influence of environmental factors on the driver's emotions, rather than treating emotions as an independent exogenous variable. In fact, after the front vehicle suddenly brakes, the driver's emotion fluctuates significantly. According to the recorded emotion data, under the fright induction condition, after the sudden braking of the front vehicle, the driver's emotion changed from a state of slight frightened $\mathrm{pad} = \{-0.168, -0.988, -0.301\}$ to a state of slight anger $\mathrm{pad} = \{-0.714, -0.273, 0.338\}$. The driver experienced rapid and violent lateral steering. Under the condition of anger induction, the driver's emotion changed from an angry state $\mathrm{pad} = \{-0.848, 0.462, 0.382\}$ to a slightly frightened state $\mathrm{pad} = \{-0.257, -0.985, -0.322\}$ after experiencing the sudden braking event of the front vehicle. The driver's driving speed significantly decreased. The structure of SOR-DBN better captures the influence of environment on emotion.

b) The DBN framework's lack of the prior knowledge of SOR experts overestimates the impact of the driver's subjective driving style on driving behavior. In the ordinary DBN, the driver's subjective driving style is regarded as an important input. In addition, the indirect influence of subjective driving style on driving behavior through emotion is also ignored, while the conflict risk and the behavior of the interactive vehicle are treated as parent nodes in SOR-DBN. Subjective driving style is an exogenous static variable that affects emotion and behavior. Figure 12 shows the conditional probability distribution of acceleration and subjective driving style inferenced by the two DBNs.

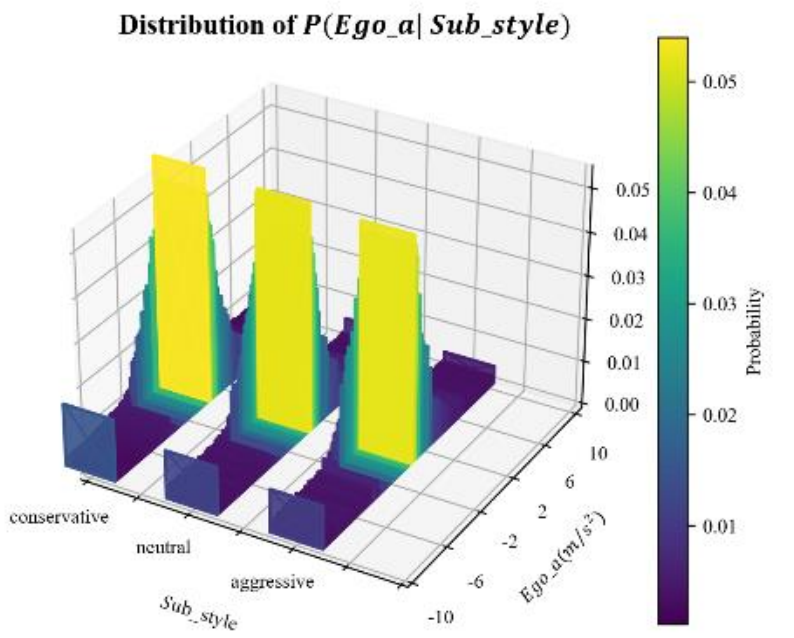

(a) Ordinary DBN

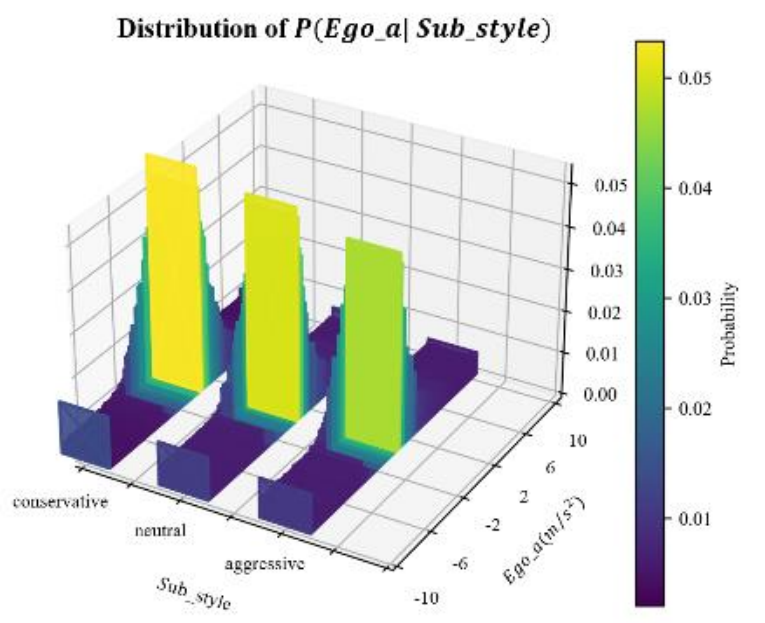

(b) SOR-DBN

**Fig. 12.** The distribution inferenced by ordinary DBN and SOR-DBN

TABLE III.
ACCURACY OF TRAJECTORY PREDICTION OF ALL METHODS

| **Scenario** | **A front car suddenly brakes** | | **A cyclist crosses the road** | | **A vehicle cuts in** | | **Unprotected left turn** | |
|---|---|---|---|---|---|---|---|---|
| **Method** | ADE | FDE | ADE | FDE | ADE | FDE | ADE | FDE |
| **GRIP++** | 2.14 | 2.84 | 0.71 | 0.92 | 3.05 | 4.22 | 0.65 | 0.81 |
| **SGCN** | 2.37 | 4.29 | 0.82 | 0.98 | 3.61 | 6.50 | 0.86 | 0.92 |
| CPSOR-GCN | **2.29** | **1.91** | **0.61** | **0.66** | **1.98** | **2.47** | **0.52** | **0.60** |

The results of DBN without the prior knowledge of SOR show that the acceleration of each subjective style driver has a similar probability distribution. Indeed, drivers with different subjective driving styles have different reactions to the varying conflict risk, and the trend of their emotion transitions also have different characteristics. These all influence the driver's behavior, resulting in different distributions of acceleration for drivers with different subjective driving styles, as reflected by SOR-DBN.

*3) Comparison Analysis*

In trajectory prediction research, graph-based deep learning models have achieved good prediction accuracy compared to other methods. The proposed CPSOR-GCN model is compared with the following State of the Art (SOTA) model.

- GRIP++[32]: Both fixed and dynamic graphs are used for trajectory predictions of different types of traffic agents. The interactive features extracted by the multi-layer graph convolution structure are input into an encoder-decoder LSTM to make predictions.
- SGCN[58]: This is a trajectory prediction model based on sparse graph convolution. Sparse directed graphs in spatial dimension are used to capture trajectory interaction features, while sparse directed graphs of time dimension are used to model motion trends. Finally, the parameters of the double Gaussian distribution for trajectory prediction are estimated by fusing the two sparse graphs mentioned above.

The prediction results are shown in Table 3. The prediction error of the proposed CPSOR-GCN model is significantly lower than that of the comparison models. Specifically, the CPSOR-GCN model we proposed is significantly better than the GRIP++ model in terms of the FDE index. The advantages are more obvious in the two long-distance trajectory predictions of the sudden braking of the front car and the vehicle cutting in. The FDE of our model has a reduction of 35% and 41% compared to the GRIP++ model. The proposed CPSOR-GCN model can better predict the mutation of vehicle trajectory features after the occurrence of emergencies. As the SOR-DBN describes the influence mechanism of emotion on driving behavior, the proposed CPSOR-GCN can better capture the features of the influence of environmental factors on emotion, as well as the variety of trajectory features induced by changes in emotion. These complex causal relationships are difficult to effectively be learned by purely data-driven deep learning models because the data often contains a lot of noise.

In addition, the GRIP++ model is based on the prediction of vehicle speed. Pre-crash scenarios usually involve drastic changes in vehicle speed. The speed features are not smooth, and the model cannot easily learn a good weight configuration. Compared with the SGCN model, the CPSOR-GCN model has more obvious advantages in prediction accuracy. The FDE of our proposed model in predicting the trajectories of the sudden braking of the front vehicle and the vehicle cutting in scenario dropped by 57% and 62%, respectively. While using sparse directional graphs to extract trajectory interaction features, SGCN uses another GCN module to simultaneously learn vehicle motion trend features in a time series. Most of its prediction results depend on the motion features of the vehicle in the previous period. However, in dangerous pre-crash scenarios, unexpected events will cause sudden changes in physical motion features. Furthermore, the driver's behavior is significantly affected by the changing emotion.

## VI. DISCUSSION AND CONCLUSIONS

In this paper, a new trajectory prediction method named CPSOR-GCN is proposed. This method simultaneously considers the driver's physical movement features and cognitive mechanism features to predict the driver's trajectory in pre-crash scenarios under abnormal emotion. The influence of abnormal emotions on driving behavior is considered based on the SOR cognitive theory. The experimental results verify the effectiveness of the SOR cognitive framework. SOR-DBN-based cognitive GCN can extract driver cognitive mechanism features more accurately. In addition, the prediction model has higher prediction accuracy than the baseline model. Especially in the two scenarios of the sudden braking of the front vehicle and the vehicle cutting in, the prediction error of the CPSOR-GCN based on the SOR-DBN cognitive framework is reduced by 36.46% and 61.58%, respectively. Compared with the SOTA model, the proposed model also has higher prediction accuracy in pre-crash scenarios. This model is very suitable for integration into active safety systems, which provides additional information to the system by accurately predicting the driving behavior under the influence of abnormal emotions to reduce the occurrence of false alarms.

Although this study has achieved certain progress and results, there are still some limitations that should be considered in future research. First, more participants will be recruited in subsequent work to expand the sample size, which will help the prediction model more comprehensively learn the mechanism of emotion on driving behavior in pre-crash scenarios. In addition, for some risk-avoiding behaviors that appear relatively low in the sample space, future work will consider giving greater learning weight to outlier samples to ensure that the model has better prediction effects for rare samples. This will provide an important direction to improve the generalization ability and adaptability of the model.

Since driver emotion recognition and calculation are not the focus of this study, a scale based on the PAD three-dimensional emotion ring model is currently used to record the driver's emotion. There is an assumption that the driver's emotion will only change once right after the sudden dangerous event occurs. In fact, the driver's emotion may fluctuate multiple times in some situations. In the future, this research can consider combining with research on emotion computing. Objective physiological data, such as facial expressions and EEG, can be used as more accurate inputs for this model to identify the driver's emotion in real-time. In addition, this study only focused on the two emotional states of fright and anger. In fact, being too happy and excited will also have a considerable influence on driving behavior. Follow-up research will be conducted on more comprehensive and more abnormal emotions.

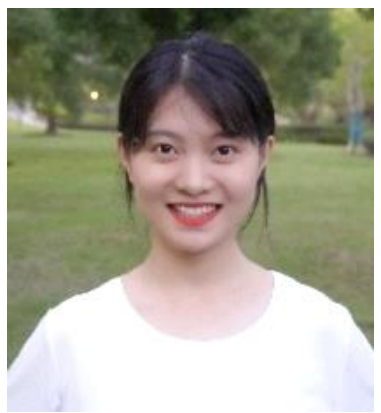

**Lanyue Tang** received the B.S. degree from southeast university, she is currently a master student at Tongji University. Her research interests include human-machine interaction, artificial intelligence, autonomous vehicle, and traffic simulation.

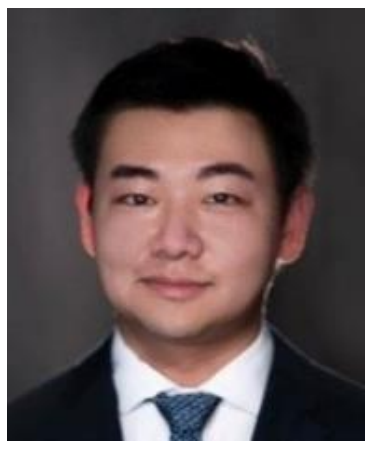

**Lishengsa Yue** (Member, IEEE) received the Ph.D. degree from the University of Central Florida. He is currently an assistant professor at Tongji University. His research relates to intelligent transportation system, human-machine interaction, driver behavior modeling and human factor. He is the handling editor of the Journal of Transportation Research Record, and he is a member in Committee on Human Factors of Vehicles of Transportation Research Board.

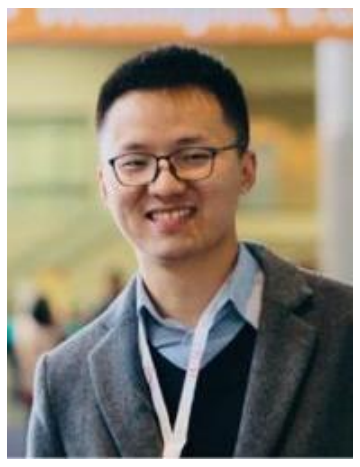

**Jinghui Yuan** received the Ph.D. degree from the University of Central Florida. He is an R&D Associate Staff member in the Applied Research for Mobility Systems (ARMS) group at the Oak Ridge National Laboratory (ORNL). With a passion for advancing transportation technologies, his research spans various areas, including intelligent transportation systems, crash risk prediction, big data analytics, deep learning, traffic simulation, driving behavior modeling, and connected and automated vehicles (CAVs).

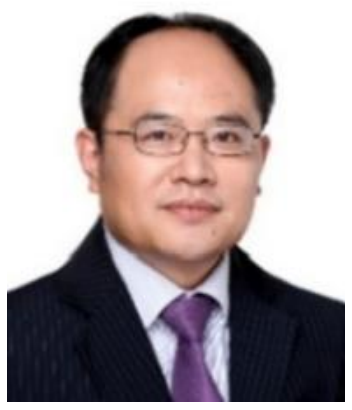

**Jian Sun** received his Ph.D. in Tongji University in 2006. Subsequently, he was at Tongji University as a Lecturer, and then promoted to the position as a Professor in 2011, where he is currently a Professor with the College of Transportation Engineering and the Dean of the Department of Traffic Engineering. His main research interests include traffic flow theory, traffic simulation, connected and automated vehicles, and intelligent transportation systems.

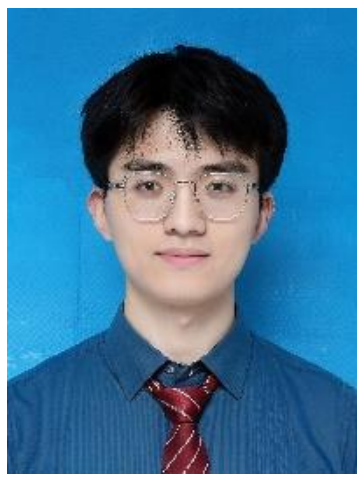

**Aohui Fu** received the B.S. degree from southeast university, he is currently a master student at Tongji University. His research interests include autonomous driving system virtual simulation, reliability verification, and autonomous driving system acceleration test.